\documentclass[twocolumn, final, sort&compress]{elsarticle}
\pdfoutput=1
\usepackage{color}
\usepackage{amsmath}
\usepackage{amssymb}
\usepackage[bookmarks=false]{hyperref}
\usepackage{epstopdf}
\usepackage{graphicx}
\usepackage{textcomp}
\usepackage{ifpdf}
\usepackage{subcaption}
\usepackage{graphicx}
\usepackage{multirow}
\usepackage[utf8]{inputenc}
\usepackage[english]{babel}
\usepackage[T1]{fontenc} 
\usepackage{array}
\usepackage{amsfonts}
\newcolumntype{C}[1]{>{\centering\let\newline\\\arraybackslash\hspace{0pt}}m{#1}}
\newcolumntype{L}[1]{>{\raggedright\let\newline\\\arraybackslash\hspace{0pt}}m{#1}}
\usepackage{mathtools}
\usepackage[makeroom]{cancel}

\usepackage{algorithm}
\usepackage{algorithmic}
\usepackage{subcaption}
\journal{Entertainment \& Computing}

\begin{document}

\begin{frontmatter}

\title{An Improved Deep Convolutional Neural Network-Based Autonomous Road Inspection Scheme Using Unmanned Aerial Vehicles}


\author[]{Syed Ali Hassan}
\author[]{Tariq Rahim}
\author[]{Soo Young Shin\corref{mycorrespondingauthor}}
\cortext[mycorrespondingauthor]{Corresponding author}
\ead{}

\address{Department of IT Convergence Engineering, Kumoh National Institute of Technology (KIT), Gumi, 39177, South Korea }
\author{syedali@kumoh.ac.kr, tariqrahim@ieee.org, wdragon@kumoh.ac.kr}


\begin{abstract}
Advancements in artificial intelligence (AI) gives a great opportunity to develop an autonomous devices. The contribution of this work is an improved convolutional neural network (CNN) model and its implementation for the detection of road cracks, potholes, and yellow lane in the road. The purpose of yellow lane detection and tracking is to realize autonomous navigation of unmanned aerial vehicle (UAV) by following yellow lane while detecting and reporting the road cracks and potholes to the server through WIFI or 5G medium. The fabrication of own data set is a hectic and time-consuming task. The data set is created, labeled and trained using default and an improved model. The performance of both these models is benchmarked with respect to accuracy, mean average precision (mAP) and detection time. In the testing phase, it was observed that the performance of the improved model is better in respect of accuracy and mAP. The improved model is implemented in UAV using the robot operating system for the autonomous detection of potholes and cracks in roads via UAV front camera vision in real-time.
\end{abstract}

\end{frontmatter}

\section{Introduction}
Deep learning (DL) which is a subset of machine learning has gained remarkable interest. It is commonly applied in facial expression recognition, self-driving cars, autonomous systems, etc [1]. Unmanned aerial vehicles (UAVs) are well-known and an attractive solution for the deployment of DL-based application. In literature, discussions regarding the deployment of DL-based algorithms are available in UAVs. moreover, the detection of UAV based on deep learning is introduced [2]. Convolutional neural network (CNN) is a deep learning-based neural network, which provides great results in object detection, especially in the case of real-time detection. A powerful neural network CNN that can be used for recognition, segmentation and detection tasks [3]. Furthermore, CNN automatically extracts features from the images and its performance is gradually improving. In the past few years, the CNN-based object detection has progressed significantly and several CNN based object detectors are introduced such as Faster R-CNN [4] and You only look once (YOLO) [5] and SSD [6] have been introduced. Specifically, there are two types of CNN-based object detectors: (1) single-shot and (2) region-based. Among these two, region-based detectors are computationally heavy and required a powerful GPU to run, whereas the single-shot detectors consist of a one convolutional neural network for detection. SSD and YOLO are types of single-shot detectors. YOLO which is developed for real-time detection, which provides excellent results [7].

Old techniques generally use  the subtraction of background [8] or different classification techniques such as Haar cascade for the detection of the objects [9]. In [10] the detection of disease in radish fields with the help of computer vision and camera attached to a drone is proposed. In [11], a convolutional neural network is utilized for the analysis of information in real-time with performance in detecting cattle with the help of a drone. In [12] autonomous computer vision based detection and landing system is proposed furthermore, in [13] the drone wireless charging method was implemented with the help of Hill-climbing algorithm. CNN based object detection is extended to medical images from the past decade for early diagnosis presented in [14].

This paper describes an improved CNN based algorithm for the autonomous inspection of roads with the improvements in the model to detect road cracks, potholes and yellow lane in real-time. Specifically, the yellow lane on the road is used as a reference for UAV to track and follow autonomously while detecting potholes and cracks on the road. YOLO is an object detector. The tiny version of YOLO is used with improvements regarding activation functions and convolutional layers.  

The paper is organized as follows: where section 2, describes related work. Section 3 describes the proposed CNN model for road cracks, potholes and yellow lane detection with yellow lane tracking is explained in details along with processes of dataset acquisition and the process of augmentation. Section 4 provides the experimental results with discussion. Finally, in section 5 the conclusion and future work of this paper is presented.

\section{RELATED WORK}
Cracks and potholes are the common road pavement defects, which are difficult to find during the inspection of road. Moreover, manual inspection of each road is difficult and costly because it requires significant effort and manpower to find cracks and potholes on time [15]. Therefore, automatic detection of cracks and potholes is introduced for reliable and speedy analysis of road defects instead of relying on the slower process of traditional manual inspection procedures [16]. Autonomous navigation of UAV using deep neural networks for indoor environments is implemented in [17]. Moreover, this navigation is utilized for outdoor environments in [18] for product delivery purposes.

The utilization of drones is increasing rapidly. In some cases, they are being operated manually through a mobile-application based joystick; whereas others are using autonomous navigation through detection and tracking of objects as implemented in [19]. UAV is also navigated using GPS and inertial navigation systems, which provide the attitude, position and velocity information that is crucial for UAV navigation, as discussed in [20]. Moreover, UAV's are being utilized to search and rescue people at sea by implementing a CNN-based person detection as implemented in [21].
\begin{figure}[h]
	\centering{\includegraphics[width=8.5cm, height=10cm]{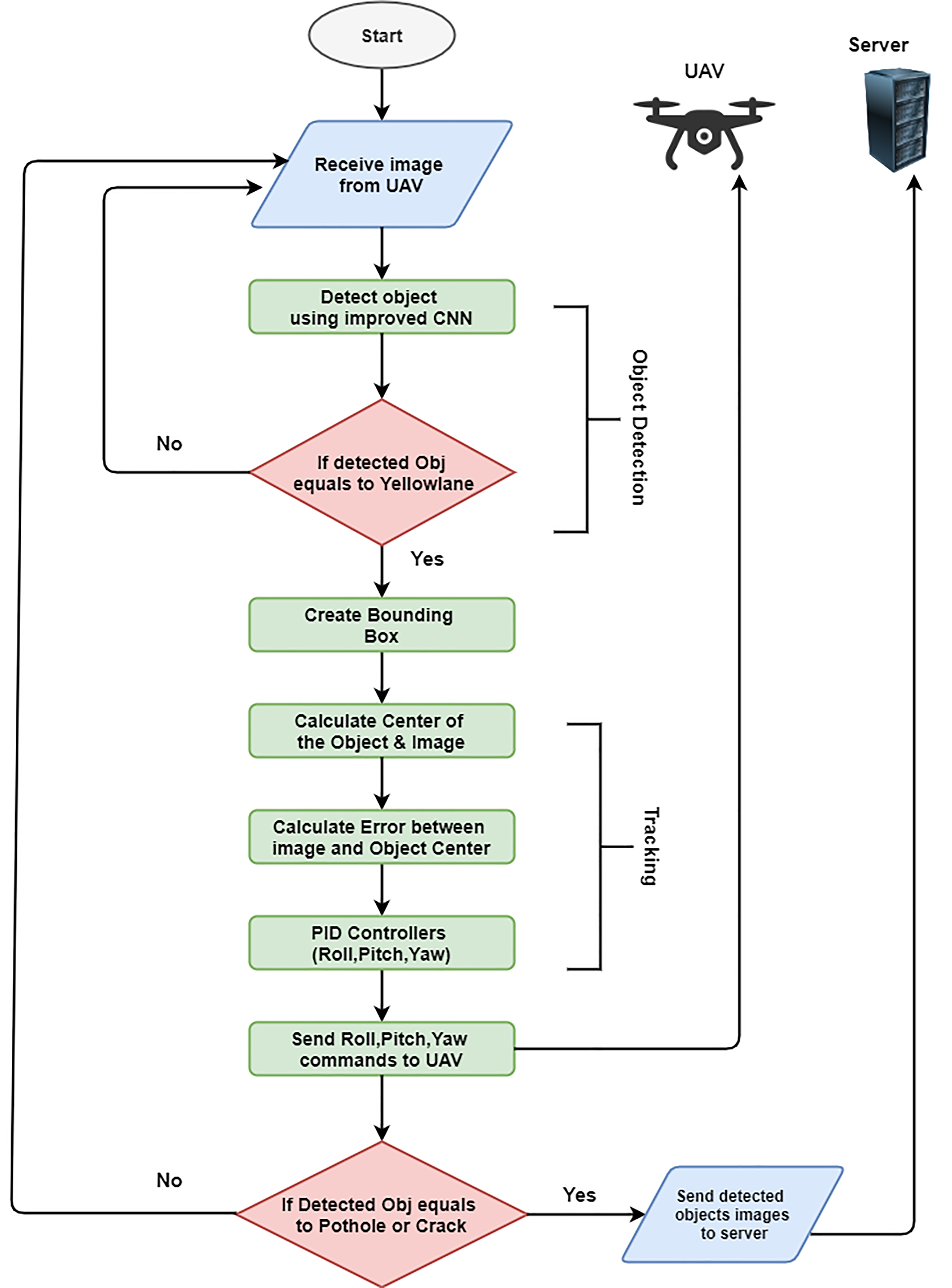}}
	\caption{Flowchart of the Proposed system.}
	\label{fig}
\end{figure}

\begin{figure}[h]
	\centering{\includegraphics[width=7.8cm, height=4cm]{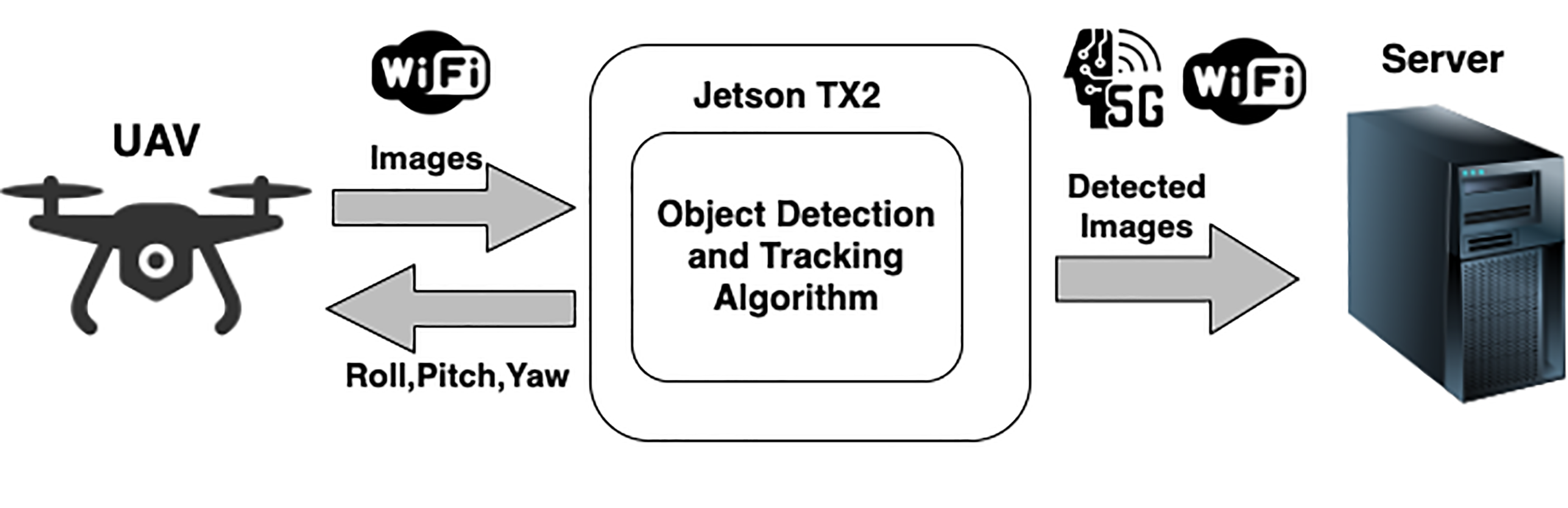}}
	\caption{Architecture of the proposed system.}
	\label{fig}
\end{figure}

\begin{figure}[h]
	\centering{\includegraphics[width=7.8cm, height=4cm]{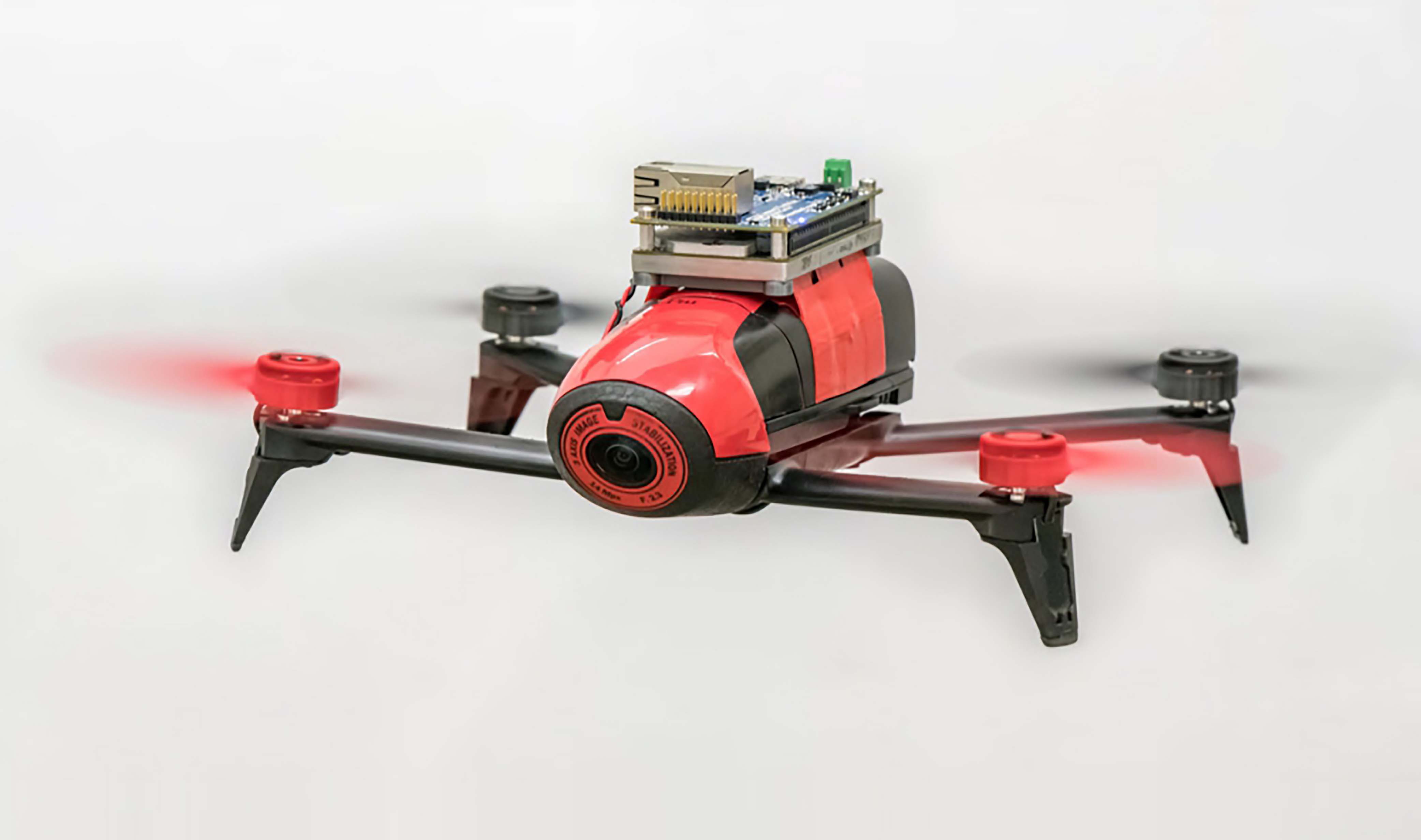}}
	\caption{The modified bebop drone 2 with tx2 mounted.}
	\label{fig}
\end{figure}
\section{Proposed Scheme}
Road defects, such as cracks and potholes are common problems that should be fixed as soon as possible. However, the inspection of roads requires sufficient manpower, and it is time-consuming. An autonomous road inspection system is proposed where a Jetson TX2 is used for the communication with Bebop drone using WIFI .  Jetson tx2 received the images from UAV using wifi. The ROS (robot operating system) is running on Jetson with the YOLO object detector. The images are received on Jetson, where ROS and YOLO are being run. If the detected object class matched with yellow lane class then the tracking would be initiated. The position and distance of the detected object are calculated to estimate pitch, roll, altitude and yaw value using the tracking algorithm. These estimated values are then sent back to guide it track and follow the yellow lane. If the detected class is identified as cracks or potholes, the detected image is sent to the server through WIFI or 5G. The flowchart of the system is illustrated in Fig 1 and the proposed architecture is depicted in Fig 2. The bebop drone is modified by integrating jetson tx2 mounted at the top of the UAV as shown in Fig 3.

\begin{figure}[t]
	\centering{\includegraphics[width=8cm, height=7cm]{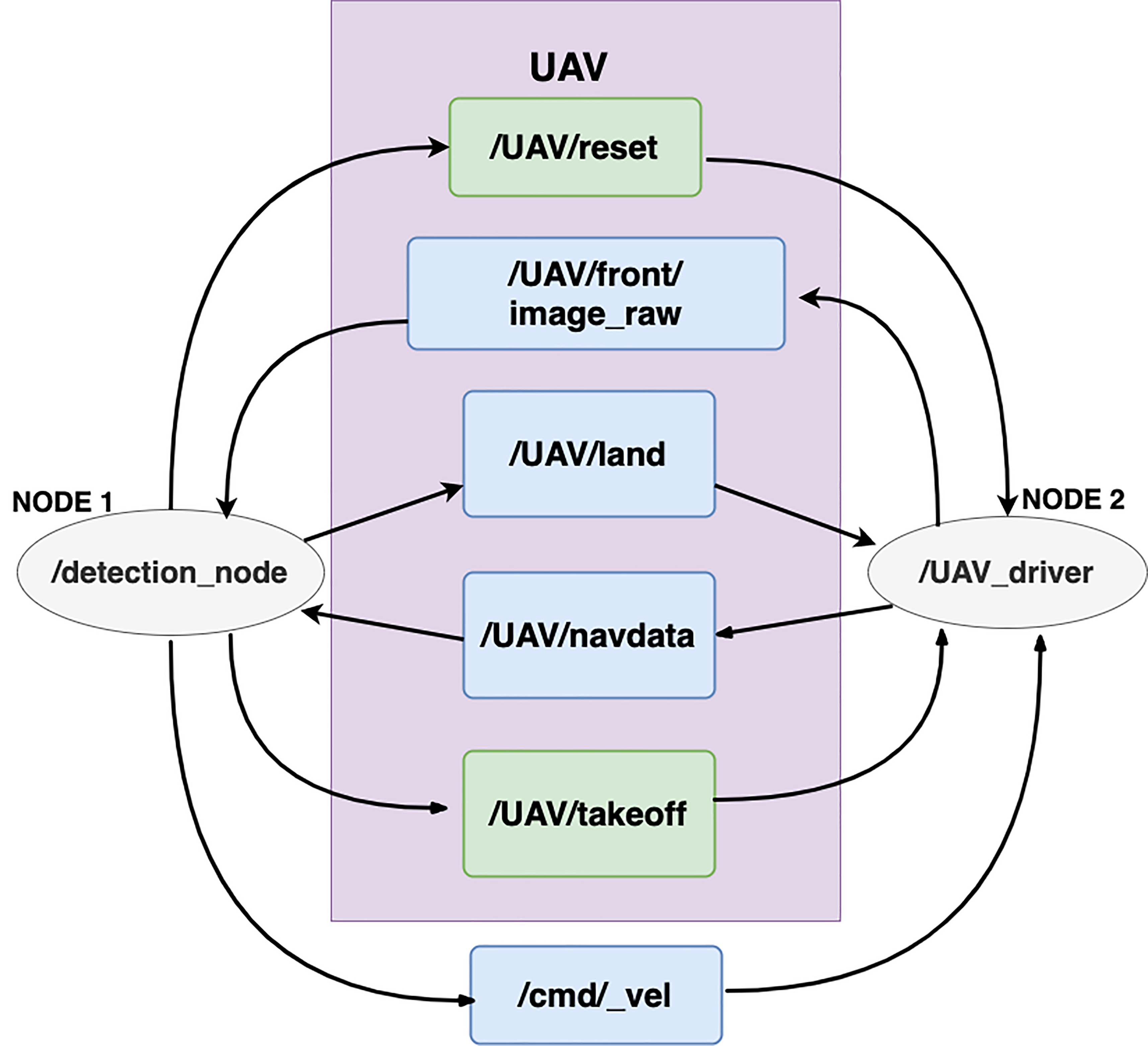}}
	\caption{ROS node graph}
	\label{fig}
\end{figure}

The tracking and object detection algorithm is implemented in UAV by utilizing Robot operating system (ROS). There are two nodes in ROS: node 01 served as  object detection and tracking  node, whereas Node 02 was the Bebop drone driver package. Both nodes communicate with each other using ROS topics [23].
In this implementation, six ROS topics were utilized. The responsibility of each topic was to carry data as a message between two nodes. The graph with nodes is illustrated in Fig 4; wherein, four ROS topics are published by Node 01. Following are the topics /\textit{UAV/reset}, \textit{/UAV/land}, \textit{/cmd\_vel} and \textit{/UAVtakeoff} . The responsibility of /cmd\_vel is to conveying the pitch, altitude (Z), yaw and roll commands to Bebop drone. Furthermore, two ROS topics are subscribed by Node 01 such as  /UAV/nav data and  /UAV/front image\_raw; which are responsible for transmitting navigation and video data, respectively. Conversely, four ROS topics, which are published by Node 01 are subscribed by Node 02 which publishes two ROS topics, namely, \textit{/UAV/navdata} and \textit{/UAV/front/image\_raw}.

In the beginning, the input image in the network is divided into a grid at the time of training phase. Bounding box "B" is predicted by each cell. There are five main characteristics of bounding box coordinates x,y at center, width and height represented as w and h, respectively; and confidence is represented as cs. The responsibility of confidence cs is to accurately determine the availability of an object in the bounding box. The detection of road cracks, pothole and a yellow lane is carried out using a lighter version of YOLO. Tiny YOLOv3 by improving the model structure. This version of YOLO is extremely fast and can be run on low-powerful devices, such as Raspberry Pi and the Jetson TX2 hardware because it has seven convolutional layers and six polling layers. However, it decreases the accuracy. Specifically, the default tiny version of YOLO employs a leaky (Relu) as an activation function and on the last layer it has linear activation function.

\begin{figure}[h]
	\centering{\includegraphics[width=8cm, height=8.5cm]{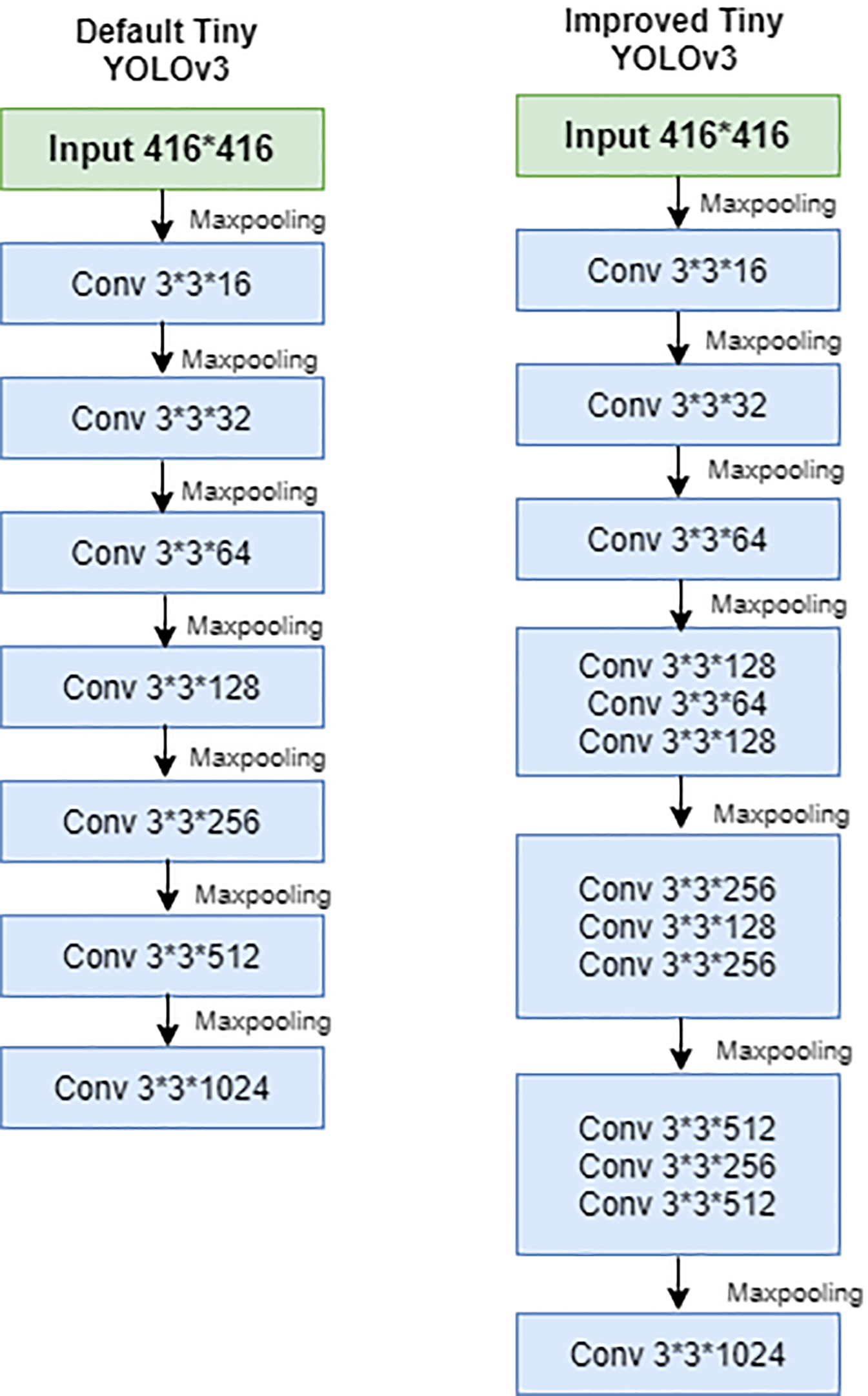}}
	\caption{Default and improved model.}
	\label{fig}
\end{figure}

\begin{figure}[h]
	\centering{\includegraphics[width=8cm, height=5cm]{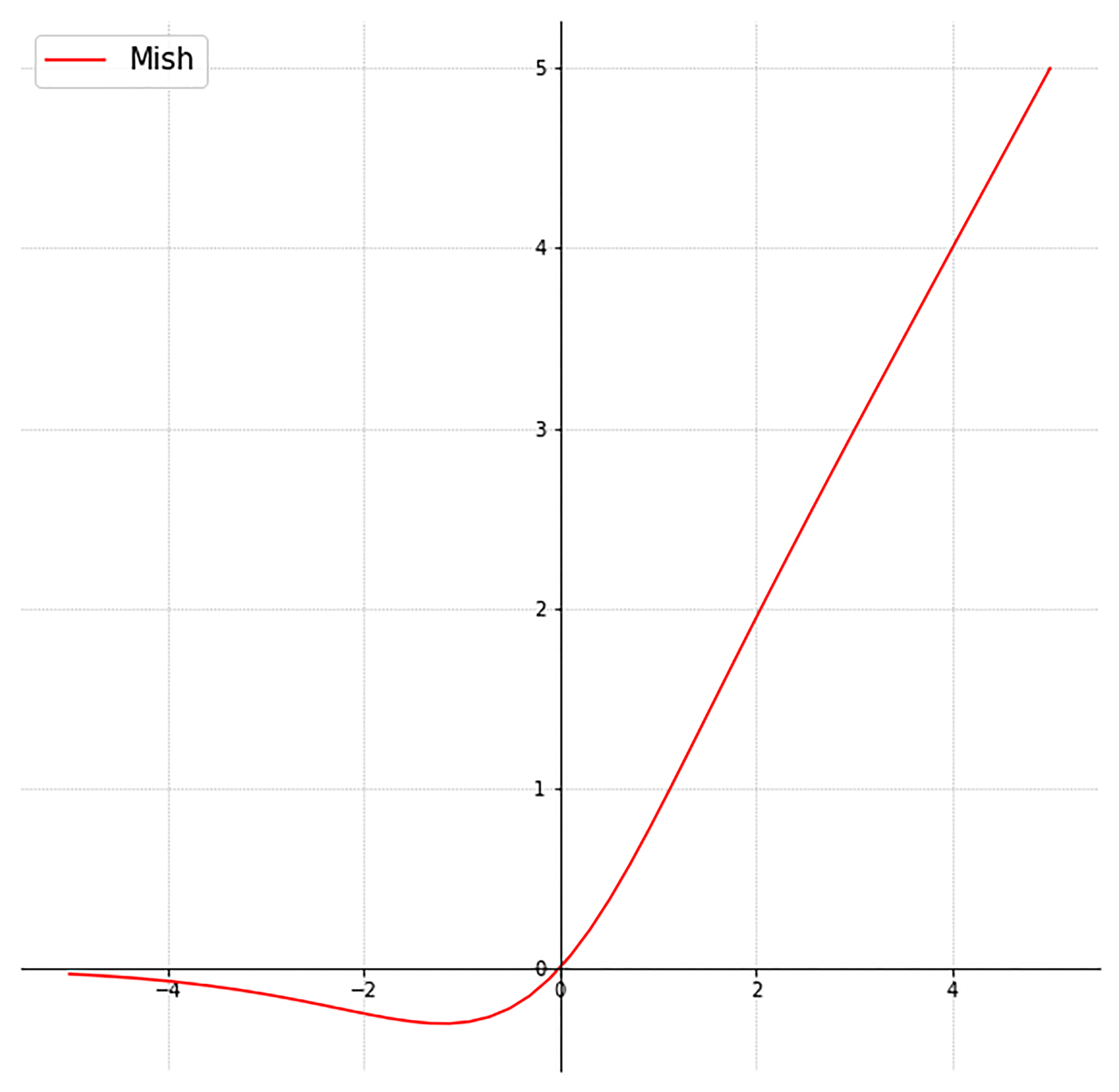}}
	\caption{Mish activation function}
	\label{fig}
\end{figure}

In the improved tiny version of YOLO, additional convolutional layers are added as shown in Fig 5, and the leaky ReLU activation function was replaced with the Mish activation function, which aids deeper propagation in hidden layers of the CNN [22] as shown in Fig 6. Mish was hence implemented to provide deeper propagation of information, better capping avoidance, and self-regularization. 

\begin{figure*}[t]
	\centering{\includegraphics[width=16cm, height=4.5cm]{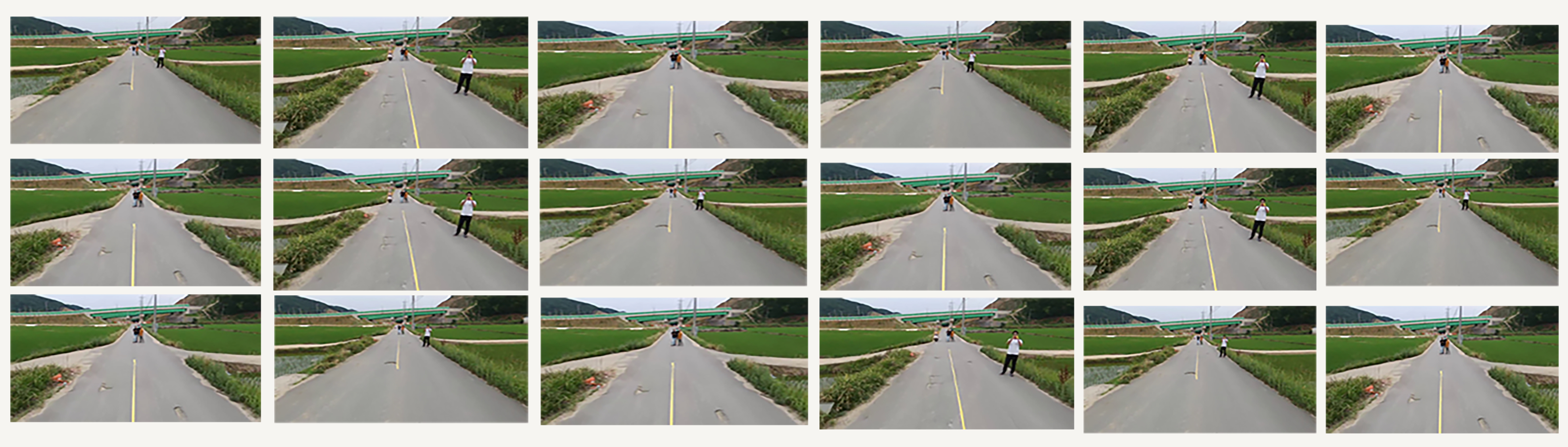}}
	\caption{Some preliminary images from the own created dataset for the implementation.}
	\label{fig}
	
\end{figure*}

\begin{figure}[t]
	\centering{\includegraphics[width=8.5cm, height=10cm]{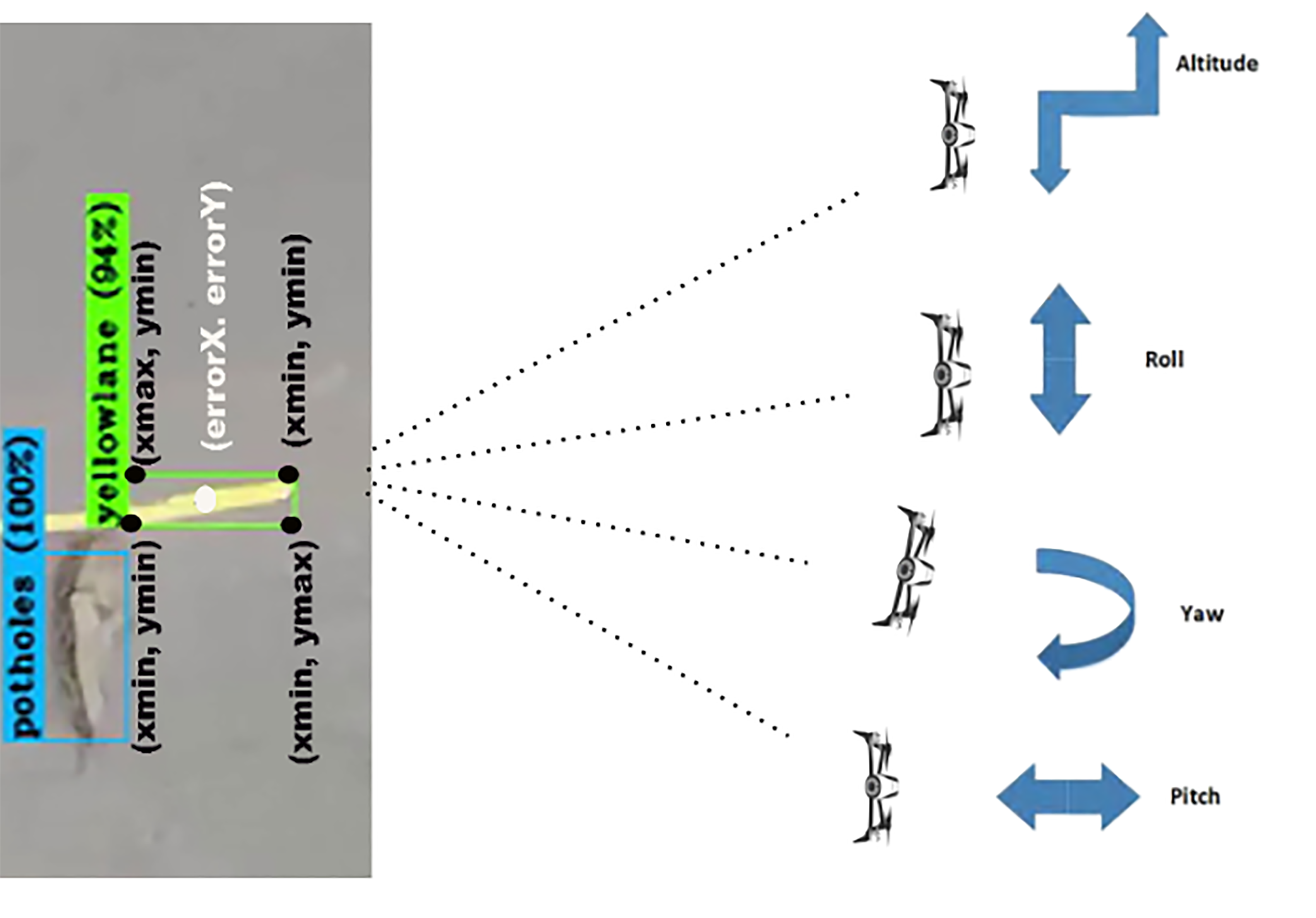}}
	\caption{Autonomous UAV Navigation}
	\label{fig}
\end{figure}
 
\begin{table}[h]
 \renewcommand{\arraystretch}{1.0}
 \caption{Parameters used in both model training}
 \resizebox{7.7cm}{!}{
  \begin{tabular}{|c|c|}
   \hline
   \textbf{Network parameters} & \textbf{Configuration values}                                                \\ \hline
   Input image dimension       & 416 X 416                                                                    \\ \hline
   Learning rate               & 0.001                                                                       \\ \hline
   Optimizer                   & \begin{tabular}[c]{@{}c@{}}Stochastic gradient \\ descent (SGD)\end{tabular} \\ \hline
   Momentum                    & 0.9                                                                          \\ \hline
   Bath size                   & 64                                                                           \\ \hline
    Subdivisions   & 4   \\ \hline
   Iterations (t)              & 10,000                                                                       \\ \hline
    Stride           & 1                                                
    \\ \hline
    Saturation           & 1.5                                              
    \\ \hline
    Exposure           & 1.5                                                
    \\ \hline
    Hue           & .1                                                
    \\ \hline
  Momentum           & 0.9                                                
    \\ \hline
  Decay          & 0.0005                                \\ \hline  
  Channels          & 3                                           \\ \hline  
  \end{tabular}
 }
\end{table}

\begin{table}[h]
 \renewcommand{\arraystretch}{1.5}
 \caption{Detection performance comparison of the proposed deep CNN model and default CNN model.}
 \resizebox{8.2cm}{!}{
  \begin{tabular}{|c|c|c|c|c|}
   \hline
   \multirow{2}{*}{\textbf{Data set}}                                                                             & \multicolumn{4}{c|}{\textbf{Performance metrics}}                                                                                      \\ \cline{2-5} 
   & \multicolumn{2}{c|}{\textbf{{Default model}} } & \multicolumn{2}{c|}{\textbf{\begin{tabular}[c]{@{}c@{}}Propose deep \\ CNN model (\%)\end{tabular}}} \\ \hline
   \multirow{5}{*}{\textbf{\begin{tabular}[c]{@{}c@{}}Cracks Class \end{tabular}}} & \textbf{Pre}                & 83.24  & \textbf{Pre}                                         & \textbf{87.63}                           \\ \cline{2-5} 
   & \textbf{Sen}                & 82.81  & \textbf{Sen}                                         & \textbf{84.02}                           \\ \cline{2-5} 
   & \textbf{F1- score}          & 83.02  & \textbf{F1- score}                                   & \textbf{85.78}                           \\ \cline{2-5} 
   & \textbf{F2- score}          & 82.89  & \textbf{F2- score}                                   & \textbf{84.71}                           \\ \cline{2-5} 
   & \textbf{Dice-coefficient}   & 83.02     & \textbf{Dice-coefficient}                            & \textbf{85.78}                           \\ \hline
  \multirow{5}{*}{\textbf{\begin{tabular}[c]{@{}c@{}}Pothole Class\end{tabular}}} & \textbf{Pre}                & 97.58.00  & \textbf{Pre}                                         & \textbf{98.26}                           \\ \cline{2-5} 
   & \textbf{Sen}                & 89.55  & \textbf{Sen}                                         & \textbf{90.12}                           \\ \cline{2-5} 
   & \textbf{F1- score}          & 93.36  & \textbf{F1- score}                                   & \textbf{94.04}                           \\ \cline{2-5} 
   & \textbf{F2- score}          & 91.04  & \textbf{F2- score}                                   & \textbf{91.63}                           \\ \cline{2-5} 
   & \textbf{Dice-coefficient}   & 88.82     & \textbf{Dice-coefficient}                            & \textbf{91.04}                           \\ \hline
   \multirow{5}{*}{\textbf{\begin{tabular}[c]{@{}c@{}}YellowLane Class\end{tabular}}}    & \textbf{Pre}                & 94.92  & \textbf{Pre}                                         & \textbf{93.26}                           \\ \cline{2-5} 
   & \textbf{Sen}                & 88.96  & \textbf{Sen}                                         & \textbf{89.45}                           \\ \cline{2-5} 
   & \textbf{F1- score}          & 91.84  & \textbf{F1- score}                                   & \textbf{91.31}                           \\ \cline{2-5} 
   & \textbf{F2- score}          & 90.09  & \textbf{F2- score}                                   & \textbf{90.10}                           \\ \cline{2-5} 
   & \textbf{Dice-coefficient}   & 91.85    & \textbf{Dice-coefficient}                            & \textbf{92.11}                            \\ \hline

  \end{tabular}
 }
\end{table}

\begin{table}[]
	\renewcommand{\arraystretch}{2.50}
	\caption{Results with comparisons of both the models}
	\centering
	\resizebox{8cm}{!}{
		\begin{tabular}{|c|c|c|} 
			\hline
			\textbf{Model} & \textbf{\begin{tabular}[c]{@{}c@{}}Mean Average Precision (mAP)\%\\ \end{tabular}} & \textbf{\begin{tabular}[c]{@{}c@{}}Accuracy\%\\ \end{tabular}} \\ \hline
			\textbf{YOLOv3Tiny}              & 85                                                                                                                                                                                                                                       & 89                                                                                   \\ \hline
			\textbf{YOLOv3TinyImproved} & 94                                                                              & 95                                                                                                                                                        \\ \hline
	\end{tabular}}
\end{table}

\section{Experimental Analysis}
This section describes the details of the own created data set and provides detailed results obtained using the improved model with the specification. Moreover, a comparison of both the default and improved model with respect to accuracy and mAP provided herein.
\begin{figure}[]
	\centering{\includegraphics[width=8cm, height=7cm]{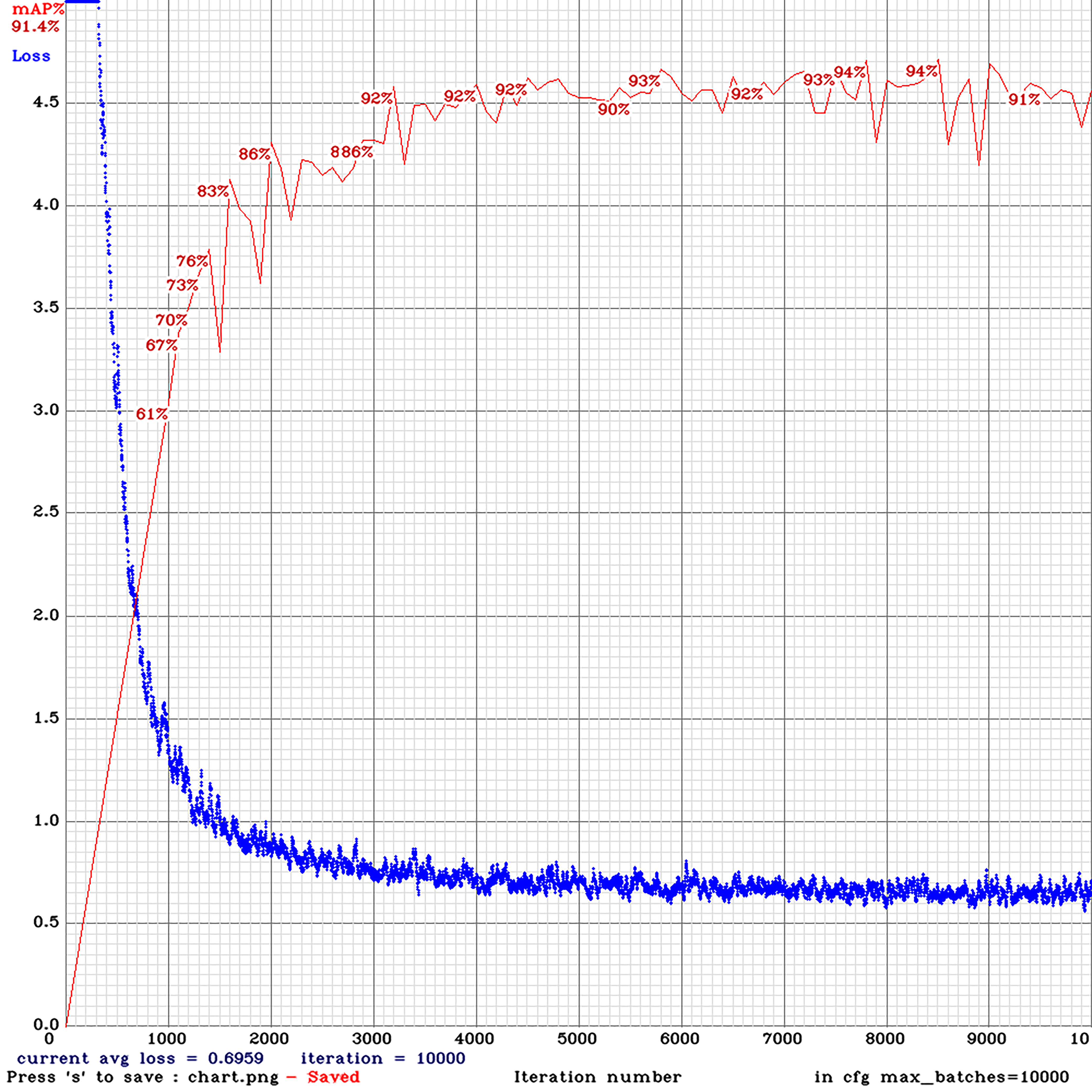}}
	\caption{Training phase of the improved model}
	\label{fig}
\end{figure}

\begin{figure}[]
	\centering{\includegraphics[width=8cm, height=8.5cm]{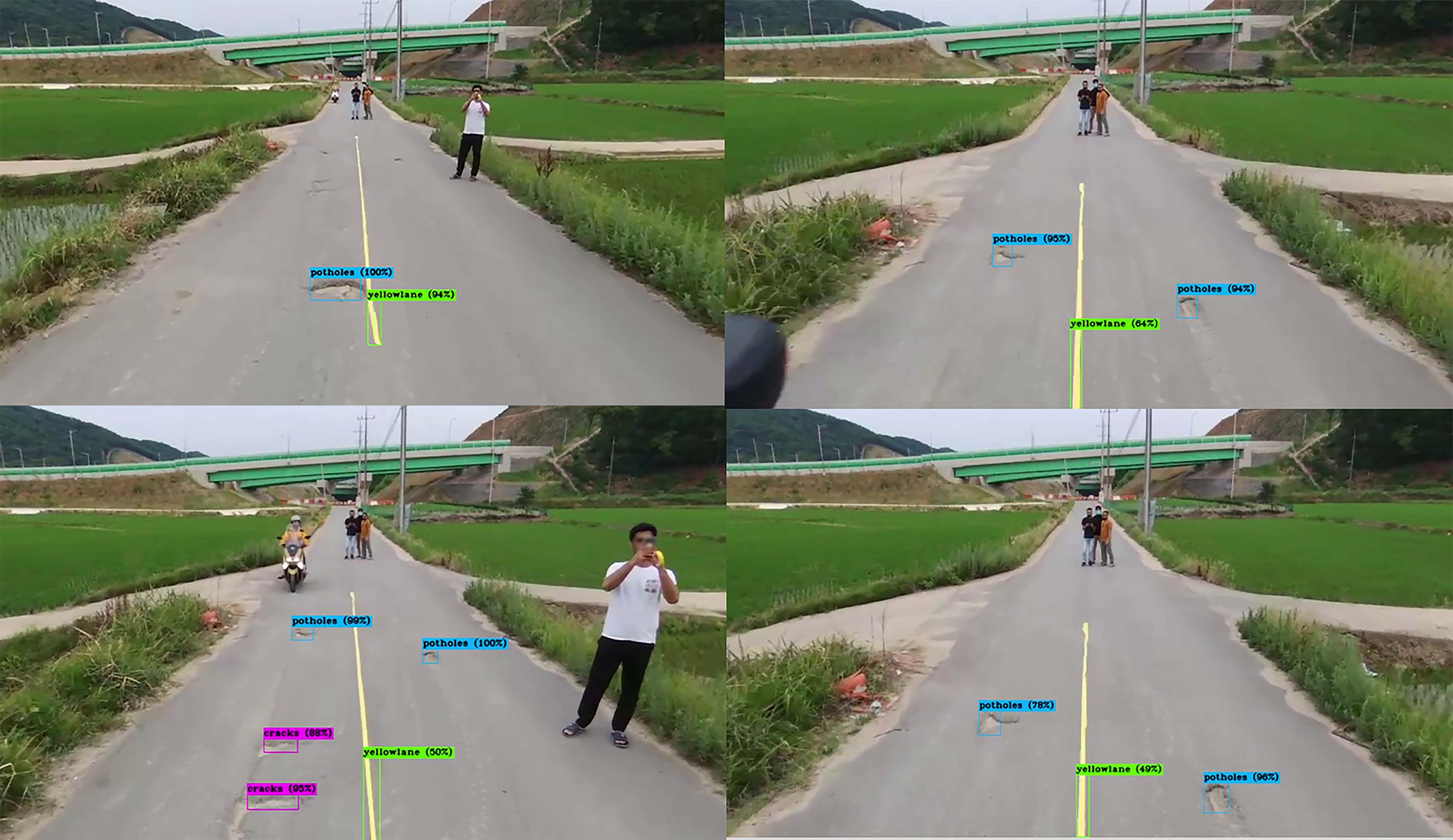}}
	\caption{Real-time results of improved model}
	\label{fig}
\end{figure}
 
\subsection{Data Set Specifications}
The creation of own data set is a difficult and hectic task. A high definition (HD) camera is used to create the own data set of road cracks, potholes and yellow lanes. The labelling of the created data set is an important task and should hence be performed carefully to achieve excellent results. The data set was divided into 80\% for training and 20\% for validation. The entire data set had three classes, namely, cracks, potholes and yellow lane which cumulatively consisted of 1000 images. The data set was first re-sized then verified after removing the bad images, and finally, improved before initiating the training. Fig 7 depicts the images used from the own created dataset for the implementation.

\subsection{Object Tacking and Navigation}
After the detection of the yellow lane in an image, the bounding box is returned by CNN. The bounding box includes the position of the object, which is represented using pixel values; (xmin,ymax),(xmax,ymax),(xmin,ymin),(xmax,ymin), as shown in Fig.8. The center of the object is calculated using these values as:

\begin{equation}
\left(x_{o}, y_{o}\right)=\left(\frac{x_{\min }+x_{\max }}{2}, \frac{y_{\min }+y_{\max }}{2}\right)
\end{equation}

The center, which is required for object tracking, can be calculated as: 

\begin{equation}
\begin{array}{l}\left(x_{i}, y_{i}\right)=\left(\frac{i m g_{w i d t h}-i m g_{w i d t h}}{2}, \frac{i m g_{h e i g h t}-i m g_{h e i g h t}}{2}\right) \\ \left(x_{i}, y_{i}\right)=(0,0)\end{array}
\end{equation}

Herein, the image center is  (0,0).  The image center and object center error is given as:

\begin{equation}
\begin{array}{l}e_{x}(t)=x_{o}-x_{i}=x_{o} \\ e_{y}(t)=y_{o}-y_{i}=y_{o}\end{array}
\end{equation}

\begin{equation}
e_{x}(t) \text { and } e_{y}(t)
\end{equation}

From the above equation, It can be concluded that for appropriately tracking the object, ex(t)  and ey(t) must always approximate or be equal to zero. Moreover, for effectively tracking, the center must be equivalent to the middle of the image in order to do tracking properly. The center of the yellow lane detected the bounding box value, which further used for tracking and following the yellow lane by the UAV with the help of roll, pitch, altitude and yaw movements as shown in Fig.9. The four control parameters are responsible for Bebop drone movement. The responsibility of roll is to move the drone left or right, for upward or downward movement pitch is responsible. The responsibility of yaw is to rotate the drone counter-clockwise or clockwise and altitude is responsible for right or left movement. The relative distance is also measured between the yellow lane and bebop drone. In order to calculate the relative distance the bounding box width of the detected yellowlane is calculated. If the bounding box width is greater than defined value, the bebop drone will go backward or it will continue its forward movement. 

\begin{figure}[h]
	\centering{\includegraphics[width=7.8cm, height=6cm]{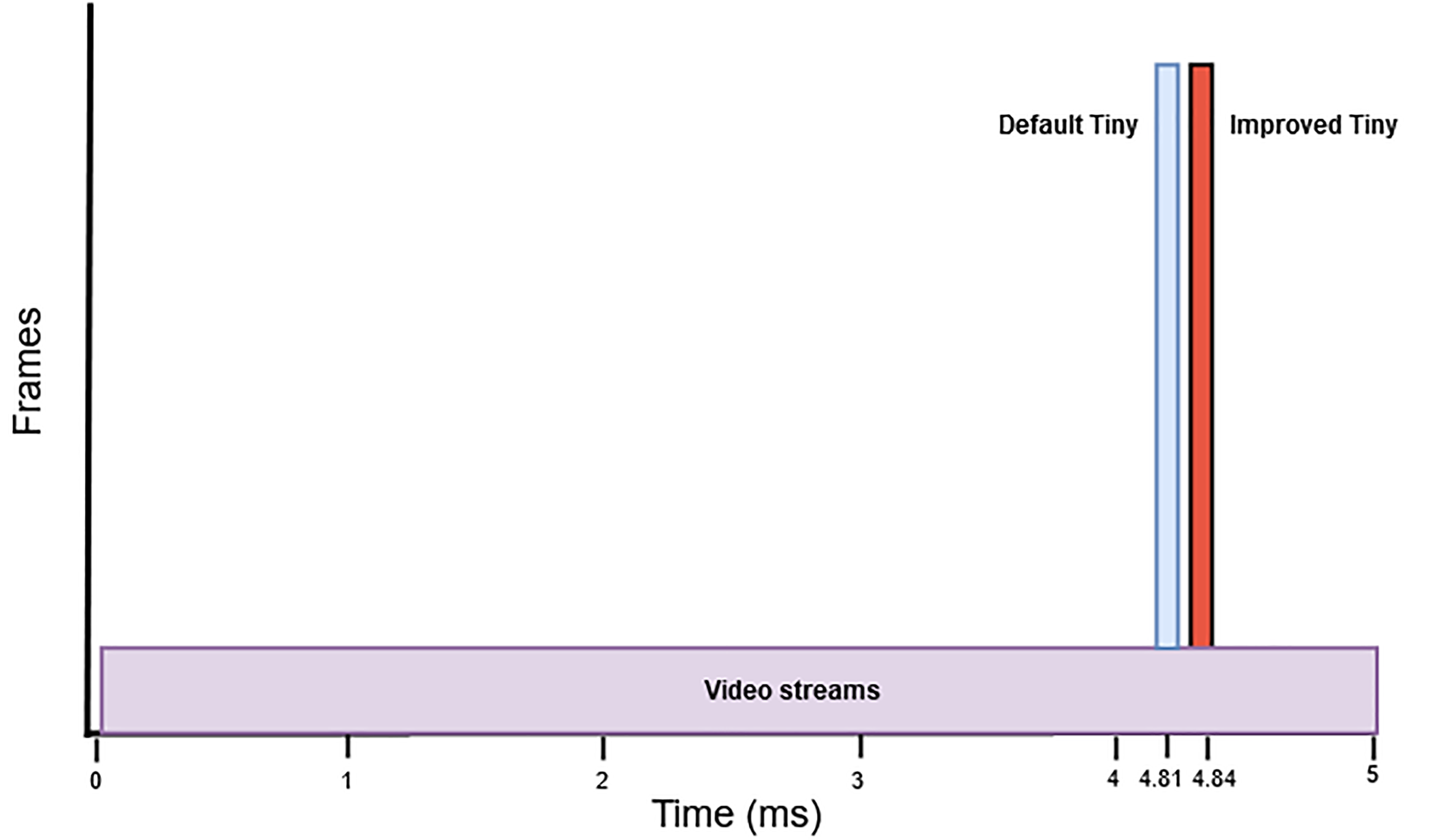}}
	\caption{Detection time comparison of both models}
	\label{fig}
\end{figure}
\subsection{Results and Training}
In the training stage, the output weights are produced after every 1000 iterations. The weight with the highest mean average precision (mAP) is considered for testing. After training was completed on the improved version of YOLO, the highest achieved MAP was found to be 94\% as shown in the real-time training chart in Fig 10. Moreover, the detection of potholes, cracks and yellow lane is shown in Fig 11. The YOLOv3 Tiny default accuracy is 89\% and its mean average precision  (mAP) was 85\% ; whereas, the improved model accuracy was 95\% which is a decent improvement. Therefore, it was observed that changing the model activation function and making the model deeper improved its accuracy. SGD is an optimizer employed for training the default and improved models with a momentum of 0.9 and learning rate of 0.001; other parameters are provided in Table 1. The improved model was trained for 10000 iterations, and after every 1000 iteration, new weight file result was produced every time. The best weight file which consists of the highest mAP was further used in the testing phase to calculate the accuracy of the model. The model is trained on a powerful GPU Titan Rtx with tensor cores allowed after 3000 iterations with a batch size of 64 and subdivision 4. The results of both models are provided in Table 2. The default model detected the object in 4.81 ms, whereas the improved model detected the object in 4.84 ms because the improved model is deeper than the default model. The comparison of detection time for both models is shown in Fig.12. 
\subsection{Performace Metrics for Evaluation}
The metrics which were used to evaluate the detection of road cracks, potholes, and yellow lane in the road, are calculated using the following parameters as defined below:

\textbf{True Positive (TP):} If the centroid falls within defined objects in the class ground truth then it is classified as true output detection. True positive is counted as one if multiple true output detection occurred within the frame.

\textbf{True Negative (TN):} The detection is true but negative frames it means frames without defined objects.

\textbf{False Positive (FP):} In this case, the detected centroid does not fall inside the defined objects in the class ground truth.

\textbf{False Negative (FN):} That is, objects that are defined in the class are missing in the frame.

In order to efficiently evaluate the performance of an improved deep CNN model by employing the parameters above.

\textbf{Precision:} This metric calculates how precisely the improved model detects the defined objects in a class within an image.

\begin{equation}
\text {Precision }(\text {Pre})=\frac{T P}{T P+F P} \times 100
\end{equation}

\textbf{Sensitivity:} This metric is also known as the true positive rate, or recall, and measures the proportion of real class of the defined object correctly.

\begin{equation}
\text {Sensitivity }(\text {Sen})=\frac{T P}{T P+F N} \times 100
\end{equation}

\textbf{F1- score and F2- score:} The harmonic mean between sensitivity and precision is known as the F1- and F2- score, and with in a range of [0,1]. To balance the sensitivity and precision, both these scores were recognized. The F1- score is given below as follows:

\begin{equation}
F 1-\text {score}=\frac{2 \times \text {Sen } \times \text { Pre}}{\text {Sen }+\text {Pre}} \times 100
\end{equation}

and the F2- score is given and calculated as:

\begin{equation}
F 2- score=\frac{5 \times \text {Pre} \times \text {Sen}}{4 \times \text {Pre}+\text {Sen}}
\end{equation}

 \textbf{Dice Coefficient:} For the comparison of the pixel-wise result between the predicted detection and ground truth that ranges [0,1], these metrics are used as follows:
 
 \begin{equation}
 \label{eq_maxminCT}
 \begin{array}{l}
 \text { Dice coefficient }(E, F)=\frac{2 \times|E
 	\cap F|}{|E|+|F|} \\
 =\frac{2 \times T P}{2 \times
 	T P+F P+F N}
 \end{array}
 \end{equation}
 
\subsection{Loss Function}
The overall process of YOLO uses the loss calculation known as a sum-square error [24]. The end to end network of YOLO which has simple differences of addition, such as coordinates errors, classifications errors and IOU errors. The below formula is used to express the loss function. 
\begin{equation}
loss = \sum\limits_{i = 0}^{{g^2}} {coordErr}  + iouErr + clsErr 
\end{equation}
The weight of each loss function is calculated to estimate the overall loss function. During the training phase, the model exhibits unstable behaviour and divergence when the classification error is constant with a coordinate error. Therefore, the coordinate error weight was fixed to $\lambda=5$. Yolo employs  $\lambda_{noobj }$ for the IOU error to keep away from confusion between the object grid and no object grid. The 
absolute loss function obtained while training the dataset can be described as follows:

\begin{equation}
\begin{array}{l}
loss = \lambda _{coord}^{}\sum\limits_{i = 0}^{g2} {} \sum\limits_{J = 0}^B {} l_{ij}^{obj}\left[ {{{\left( {{a_i} - {{\hat a}_i}} \right)}^2} + {{\left( {{b_i} - {{\hat b}_i}} \right)}^2}} \right]\\
 + {\lambda _{coord}}\sum\limits_{i = 0}^{g2} {} \sum\limits_{j = 0}^B {} l_{ij}^{obj}\\
 \left[ {{{\left( {\sqrt {{w_i}}  - \sqrt {{{\hat w}_i}} } \right)}^2} + {{\left( {\sqrt {{h_i}}  - \sqrt {{{\hat h}_i}} } \right)}^2}} \right]\\
 + \sum\limits_{i = 0}^{g2} {} \sum\limits_{j = 0}^B {} l_{ij}^{obj}{\left( {{c_i} - {{\hat c}_i}} \right)^2}\\
 + {\lambda _{noobj}}\sum\limits_{i = 0}^{g2} {} \sum\limits_{j = 0}^B {} l_{ij}^{obj}{\left( {{c_i} - {{\hat c}_i}} \right)^2}\\
 + \sum\limits_{i = 0}^{g2} {} l_i^{obj}\sum\limits_{c \in class}^{} {} {\left( {{R_i}(c) - {{\hat R}_i}(c)} \right)^2}
\end{array}
\end{equation}

In the above equation, the number of grids is represented by g. Each of the cell numbers corresponding to the prediction boxes are indicated using B. The coordinate center of each cell is defined as (a,b); moreover, its width and height are indicated as h,w, respectively. Furthermore, the prediction box confidence is indicated as C; the confidence of objects in the class is labelled as R. The weight of the loss function position is represented as $\lambda_{coord }$. The classification loss function weight is defined as $\lambda_{noobj }$ If the objects that are trained in this class are present then the value is set as 1, and otherwise it is 0.

\section{CONCLUSION AND FUTURE WORK:}

\label{sec:7}

A convolutional neural network was improved and implemented for the detection of cracks, potholes and yellow lane in the road. Autonomous navigation of UAV is achieved by tracking and following the yellow lane for road inspection to report road damages on the server. An HD data set was produced to achieve the best results. Subsequently, the results obtained from both models were compared in terms of detection time, mAP and accuracy.
The future work will be conducted after creating a larger dataset and compare its results with those of other object detectors such as Faster-RCNN.

\section*{Acknowledgment}

This work was supported by Priority Research Centers Program through the National Research Foundation of Korea (NRF) funded by the Ministry of Education, Science and Technology(2018R1A6A1A03024003)


\begin{thebibliography}{1}
\expandafter\ifx\csname url\endcsname\relax
  \def\url#1{\texttt{#1}}\fi
\expandafter\ifx\csname urlprefix\endcsname\relax\def\urlprefix{URL }\fi
\expandafter\ifx\csname href\endcsname\relax
  \def\href#1#2{#2} \def\path#1{#1}\fi

\bibitem{Feynman1963118}
R.~Feynman, F.~{Vernon Jr.}, The theory of a general quantum system interacting
  with a linear dissipative system, Annals of Physics 24 (1963) 118--173.
\newblock \href {http://dx.doi.org/10.1016/0003-4916(63)90068-X}
  {\path{doi:10.1016/0003-4916(63)90068-X}}.

\bibitem{Dirac1953888}
P.~Dirac, The lorentz transformation and absolute time, Physica 19~(1-–12)
  (1953) 888--896.
\newblock \href {http://dx.doi.org/10.1016/S0031-8914(53)80099-6}
  {\path{doi:10.1016/S0031-8914(53)80099-6}}.

\end{thebibliography}


\begin{thebibliography}{}
\bibitem{IEEEhowto:kopka}
Mantoro, Teddy, and Media A. Ayu. "Multi-faces recognition process using Haar cascades and eigenface methods." In 2018 6th International Conference on Multimedia Computing and Systems (ICMCS), pp. 1-5. IEEE, 2018.

\bibitem{IEEEhowto:kopka}
Hassan, Syed Ali, Tariq Rahim, and Soo Young Shin. "Real-time UAV Detection based on Deep Learning Network." In 2019 International Conference on Information and Communication Technology Convergence (ICTC), pp. 630-632. IEEE, 2019.

\bibitem{IEEEhowto:kopka}
Audebert, Nicolas, Bertrand Le Saux, and SÃbastien LefÃ¨vre. "Segment-before-detect: Vehicle detection and classification through semantic segmentation of aerial images." Remote Sensing 9, no. 4 (2017): 368.
\bibitem{IEEEhowto:kopka}
Ren, Shaoqing, Kaiming He, Ross Girshick, and Jian Sun. "Faster r-cnn: Towards real-time object detection with region proposal networks." In Advances in neural information processing systems, pp. 91-99. 2015.

\bibitem{IEEEhowto:kopka}
Redmon, Joseph, Santosh Divvala, Ross Girshick, and Ali Farhadi. "You only look once: Unified, real-time object detection." In Proceedings of the IEEE conference on computer vision and pattern recognition, pp. 779-788. 2016.

\bibitem{IEEEhowto:kopka}
Liu, Wei, Dragomir Anguelov, Dumitru Erhan, Christian Szegedy, Scott Reed, Cheng-Yang Fu, and Alexander C. Berg. "Ssd: Single shot multibox detector." In European conference on computer vision, pp. 21-37. Springer, Cham, 2016.

\bibitem{IEEEhowto:kopka}
J. Redmon, Darknet: Open Source Neural Networks in C, 2016, [online] Available: http://pjreddie.com/darknet/.

\bibitem{IEEEhowto:kopka}
Philipps, Jonathan Johannes, Ingrid BAnninger, Martin Weigert, and Javier VÃ¡squez. "Automatic tracking and counting of moving objects." In 3rd IEEE International Work-Conference on Bioinspired Intelligence, pp. 93-97. IEEE, 2014.

\bibitem{IEEEhowto:kopka}
Budiman, Rezha Aditya Maulana, Balza Achmad, Agus Arif, and Luthfi Zharif. "Localization of white blood cell images using Haar cascade classifiers." In 2016 1st International Conference on Biomedical Engineering (IBIOMED), pp. 1-5. IEEE, 2016.

\bibitem{IEEEhowto:kopka}
Dang, L. Minh, Syed Ibrahim Hassan, Im Suhyeon, Arun kumar Sangaiah, Irfan Mehmood, Seungmin Rho, Sanghyun Seo, and Hyeonjoon Moon. "UAV based wilt detection system via convolutional neural networks." Sustainable Computing: Informatics and Systems (2018).

\bibitem{IEEEhowto:kopka}
Rivas, Alberto, Pablo Chamoso, Alfonso GonzÃ¡lez-Briones, and Juan Manuel Corchado. "Detection of cattle using drones and convolutional neural networks." Sensors 18, no. 7 (2018): 2048.

\bibitem{IEEEhowto:kopka}
Rabah, Mohammed, Ali Rohan, Muhammad Talha, Kang-Hyun Nam, and Sung Ho Kim. "Autonomous vision-based target detection and safe landing for UAV." International Journal of Control, Automation and Systems 16, no. 6 (2018): 3013-3025.

\bibitem{IEEEhowto:kopka}
Rohan, Ali, Mohammed Rabah, Furqan Asghar, Muhammad Talha, and Sung-Ho Kim. "Advanced drone battery charging system." Journal of Electrical Engineering and Technology 14, no. 3 (2019): 1395-1405.

\bibitem{IEEEhowto:kopka}
Rahim, Tariq, Muhammad Arslan Usman, and Soo Young Shin. "A Survey on Contemporary Computer-Aided Tumor, Polyp, and Ulcer Detection Methods in Wireless Capsule Endoscopy Imaging." arXiv preprint arXiv:1910.00265 (2019).

\bibitem{IEEEhowto:kopka}
Cafiso, Salvatore, A. D. Graziano, and Sebastiano Battiato. "Evaluation of pavement surface distress using digital image collection and analysis." In Seventh International Congress on Advances in Civil Engineering, pp. 1-10. 2006.

\bibitem{IEEEhowto:kopka}
Huang, Yaxiong, and Bugao Xu. "Automatic inspection of pavement cracking distress." Journal of Electronic Imaging 15, no. 1 (2006): 013017.

\bibitem{IEEEhowto:kopka}
Padhy, Ram Prasad, Sachin Verma, Shahzad Ahmad, Suman Kumar Choudhury, and Pankaj Kumar Sa. "Deep neural network for autonomous uav navigation in indoor corridor environments." Procedia computer science 133 (2018): 643-650.

\bibitem{IEEEhowto:kopka}
MuÃ±oz, Guillem, Cristina Barrado, Ender Ã‡etin, and Esther Salami. "Deep reinforcement learning for drone delivery." Drones 3, no. 3 (2019).

\bibitem{IEEEhowto:kopka}
Boudjit, K., and Chevif Larbes. "Detection and implementation autonomous target tracking with a Quadrotor AR. Drone." In 2015 12th International Conference on Informatics in Control, Automation and Robotics (ICINCO), vol. 2, pp. 223-230. IEEE, 2015.

\bibitem{IEEEhowto:kopka}
Ding, Weidong, Jinling Wang, and Ali Almagbile. "Adaptive filter design for UAV navigation with GPS/INS/optic flow integration." In 2010 International Conference on Electrical and Control Engineering, pp. 4623-4626. IEEE, 2010.

\bibitem{IEEEhowto:kopka}
Wang, Shubo, Yu Han, Jian Chen, Zichao Zhang, Guangqi Wang, and Nannan Du. "A Deep-Learning-Based Sea Search and Rescue Algorithm by UAV Remote Sensing." In 2018 IEEE CSAA Guidance, Navigation and Control Conference (CGNCC), pp. 1-5. IEEE, 2018.
\bibitem{IEEEhowto:kopka}
Misra, Diganta. "Mish: A self regularized non-monotonic neural activation function." arXiv preprint arXiv:1908.08681 (2019).
\bibitem{IEEEhowto:kopka}
Quigley, Morgan, Ken Conley, Brian Gerkey, Josh Faust, Tully Foote, Jeremy Leibs, Rob Wheeler, and Andrew Y. Ng. "ROS: an open-source Robot Operating System." In ICRA workshop on open source software, vol. 3, no. 3.2, p. 5. 2009.
\bibitem{IEEEhowto:kopka}
Ranjbar, Mani, Greg Mori, and Yang Wang. Optimizing complex
loss functions in structured prediction. In European Conference on
Computer Vision, pp. 580-593. Springer, Berlin, Heidelberg, 2010.

\end{thebibliography}
\end{document}